\documentclass[runningheads]{llncs}

\usepackage{eccv}

\usepackage{multirow}
\usepackage[table]{xcolor}

\usepackage{color}
\usepackage{soul}

\usepackage{eccvabbrv}

\usepackage{graphicx}
\usepackage{booktabs}

\usepackage[accsupp]{axessibility}  

\usepackage[pagebackref,breaklinks,colorlinks,citecolor=eccvblue]{hyperref}

\usepackage{orcidlink}

\begin{document}

\title{HiVLA: A Visual-Grounded-Centric Hierarchical Embodied Manipulation System} 

\titlerunning{HiVLA}


\author{Tianshuo Yang\inst{1,2^*}, Guanyu Chen\inst{3^*}, Yutian Chen\inst{2,4}, Zhixuan Liang\inst{1,2},\\ Yitian Liu\inst{3}, Zanxin Chen\inst{2,3}, Chunpu Xu\inst{2}, Haotian Liang\inst{2}, Jiangmiao Pang\inst{2}, Yao Mu\inst{3,2\dagger}, Ping Luo\inst{1\dagger}
}
\renewcommand{\thefootnote}{\fnsymbol{footnote}}
\footnotetext[0]{\hspace{-1.1em}\textsuperscript{*} Equal contribution \textsuperscript{\dag} Corresponding authors:
\href{mailto:muyao@sjtu.edu.cn}{muyao@sjtu.edu.cn},
\href{mailto:pluo@cs.hku.hk}{pluo@cs.hku.hk}}

\authorrunning{T.~Yang et al.}


\institute{
The University of Hong Kong \and
Shanghai AI Laboratory \and
Shanghai Jiao Tong University \and
The Chinese University of Hong Kong
}

\maketitle

\begin{abstract}
While end-to-end Vision-Language-Action (VLA) models offer a promising paradigm for robotic manipulation, fine-tuning them on narrow control data often compromises the profound reasoning capabilities inherited from their base Vision-Language Models (VLMs). To resolve this fundamental trade-off, we propose HiVLA, a visual-grounded-centric hierarchical framework that explicitly decouples high-level semantic planning from low-level motor control. In high-level part, a VLM planner first performs task decomposition and visual grounding to generate structured plans, comprising a subtask instruction and a precise target bounding box. Then, to translate this plan into physical actions, we introduce a flow-matching Diffusion Transformer (DiT) action expert in low-level part equipped with a novel cascaded cross-attention mechanism. This design sequentially fuses global context, high-resolution object-centric crops and skill semantics, enabling the DiT to focus purely on robust execution. Our decoupled architecture preserves the VLM's zero-shot reasoning while allowing independent improvement of both components. Extensive experiments in simulation and the real world demonstrate that HiVLA significantly outperforms state-of-the-art end-to-end baselines, particularly excelling in long-horizon skill composition and the fine-grained manipulation of small objects in cluttered scenes. The project website is: \href{https://tianshuoy.github.io/HiVLA-page/}{https://tianshuoy.github.io/HiVLA-page/}
\keywords{Vision-Language-Action Models \and VLM Agent Systems}

\end{abstract}
    
\section{Introduction}
\label{sec:intro}


Achieving human-like capabilities in robots that integrates perception, reasoning and execution, is a central pursuit of embodied AI. Recently, the advent of web-scale, pretrained Vision-Language Models~\cite{bai2025qwen3,steiner2024paligemma,liu2023improved} (VLMs) has presented a transformative opportunity for robotic manipulation. Exhibiting remarkable zero-shot generalization and deep semantic understanding, VLMs have catalyzed the development of Vision-Language-Action (VLA) models~\cite{kim2025openvla,black2024pi_0,intelligence2025pi05visionlanguageactionmodelopenworld}. However, current VLA research predominantly adopts end-to-end architectures, utilizing either single-system~\cite{brohan2023rt2,zhao2025cot,kim2025fine} or dual-system~\cite{bu2024towards,bjorck2025gr00t,cheang2025gr3} approaches that tightly couple visual reasoning with low-level action generation. Although these integrated paradigms have shown considerable promise, they face a critical bottleneck~\cite{hancock2025actionslanguagefinetuningvlms,driess2025knowledgeinsulatingvisionlanguageactionmodels} that fine-tuning VLMs on relatively scarce and domain-specific manipulation data inevitably degrades their original reasoning capabilities. This degradation, widely recognized as catastrophic forgetting, ultimately limits the ability to leverage the full cognitive power of the most advanced VLMs.

\begin{figure}[tb]
    \centering
    \includegraphics[width=1.0\textwidth]{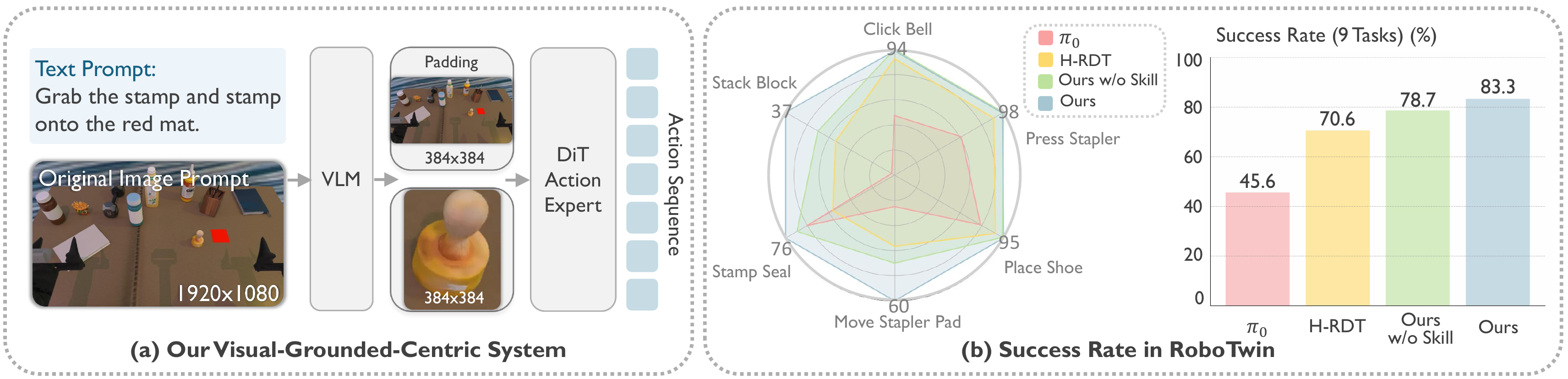}
    \vspace{-10pt}
    \caption{
        (a) Overview of our proposed HiVLA framework. (b) Success rate comparison on RoboTwin benchmark.
    }
    \label{fig:teaser}
    \vspace{-10pt}
\end{figure}

Hierarchical systems~\cite{belkhale2024rt,shi2025hirobotopenendedinstruction,liang2024skilldiffuserinterpretablehierarchicalplanning} offer a compelling alternative by explicitly decoupling high-level semantic planning from low-level motor control. In this paradigm, the VLM operates purely as a high-level planner, preserving its reasoning capabilities by avoiding low-level fine-tuning, while a dedicated action expert executes the plans. However, the success of this decoupled design heavily depends on the intermediate representation bridging the two modules. A powerful candidate for this interface is visual grounding. This concept is deeply inspired by the ``thinking with images'' paradigm in VLM agents~\cite{Jiang_2025_ICCV,Shi_2025_CVPR,lai2025minio3scalingreasoningpatterns,zheng2025deepeyesincentivizingthinkingimages}, a framework in which a model explicitly localizes a relevant target region in a high-resolution image before proceeding with complex reasoning.


Despite the conceptual elegance of visual-grounded-centric VLAs, existing designs struggle to effectively translate grounded information into physical actions. Current methods typically force a compromise between spatial context and visual fidelity. For example, extracting local image crops often strips away absolute spatial coordinates~\cite{fan2025interleave}. Conversely, applying object masks to down-sampled global images discards the nuanced visual details necessary for fine-grained manipulation~\cite{zhong2025dexgraspvlavisionlanguageactionframeworkgeneral}.These shortcomings expose a critical, unresolved question that how can we design a policy capable of fully exploiting a grounded plan including high-resolution local appearance, precise global spatial awareness, and explicit skill-level subtask directives?

To address this challenge, we propose HiVLA, a hierarchical manipulation system centered around a robust framework for visual-grounded plan generation and utilization. As illustrated in~\cref{fig:teaser} (a), our system employs a VLM as a high-level planner that decomposes complex instructions and visually grounds target objects. This process outputs a structured plan consisting of a semantic subtask label and a precise bounding box. To effectively translate this fine-grained guidance into physical motion, we design a low-level action expert based on a Diffusion Transformer (DiT)~\cite{peebles2023scalable}. Within this expert, our key innovation is a cascaded cross-attention mechanism embedded in each DiT block. Rather than naively fusing inputs, this mechanism sequentially conditions the policy on three distinct signals: \emph{(1)} global visual context for holistic scene understanding, \emph{(2)} high-resolution, object-centric features from the grounded patch augmented with absolute positional encodings to preserve spatial awareness, and \emph{(3)} a language embedding representing the specific subtask skill. This architectural grounding design enables the action expert to maximally leverage the VLM's cognitive output, providing the system with a clear understanding of what to do, where to look, and how to act.

Experiments conducted in two challenging, cluttered simulation environments and the real world demonstrate the superiority of our approach. As shown in~\cref{fig:teaser} (b), HiVLA achieves an absolute success rate improvement of 12.7\% over a strong baseline H-RDT~\cite{bi2025h} and 37.7\% over the state-of-the-art $\pi_0$~\cite{black2024pi_0} on the RoboTwin 2.0 Benchmark~\cite{chen2025robotwin}. These results validate that our visual-grounded-centric hierarchy significantly enhances robust perception, precise manipulation, and long-horizon task completion.
Our contributions are summarized as follows:

\begin{itemize}
\item We propose HiVLA, a hierarchical VLA framework bridged by a visual-grounded-centric mechanism. This architecture explicitly decouples VLM-based high-level planning from low-level control, eliminating catastrophic forgetting of multi-task manipulation and allowing seperate improvements of VLM and action expert.
\item We introduce a novel cascaded cross-attention mechanism within the DiT action expert, capable of sequentially integrating global context, spatially-aware high-resolution local crops, and subtask skill guidance, unlocking the potential of grounded plans for precise action generation.
\item We perform extensive evaluations in simulation and the real world, demonstratingHiVLA significantly outperforms state-of-the-art VLA models, and showcases exceptional proficiency in long-horizon skill composition and fine-grained manipulation within highly cluttered environments.
\end{itemize}

\section{Related Work}
\label{sec:related}
\subsection{Vision-Language-Action Models}
Vision-Language-Action (VLA) models have revolutionized robotic manipulation by leveraging the profound cognitive abilities of large Vision-Language Models (VLMs) to translate multi-modal inputs into executable actions. Current monolithic VLA architectures broadly fall into single-system and dual-system paradigms \cite{shao2025largevlm}. Single-system models, such as RT-2~\cite{brohan2023rt2} and OpenVLA~\cite{kim2025openvla}, employ a unified network that directly decodes action tokens autoregressively from sensory inputs. Alternatively, dual-system models like $\pi_0$\cite{black2024pi_0} and GR00T-N1.5\cite{bjorck2025gr00t} utilize a VLM backbone to implicitly guide an action expert through jointly optimized feature spaces. Although these integrated approaches demonstrate significant promise, fine-tuning VLMs on narrow manipulation data severely degrades their original, web-scale reasoning capabilities~\cite{hancock2025actionslanguagefinetuningvlms,driess2025knowledgeinsulatingvisionlanguageactionmodels}. This catastrophic forgetting limits the generalization potential of the underlying foundation models.

To circumvent this limitation, hierarchical models explicitly decouple high-level task planning from low-level policy execution via interpretable intermediate representations. This modularity retains the VLM's zero-shot reasoning power while allowing the action expert to specialize in precise motor control. These intermediate bridges take various forms, including textual subtasks in HiRobot \cite{shi2025hirobotopenendedinstruction}and MemER~\cite{sridhar2025memer} or spatial keypoints in HAMSTER \cite{li2025hamsterhierarchicalactionmodels}. By isolating cognitive processes from high-frequency control, hierarchical systems provide a robust and scalable foundation for advancing embodied intelligence.

\subsection{Visual-Grounded-Centric VLA}
A critical challenge in manipulation is precise visual grounding, which accurately maps high-level instructions to specific spatial regions within the visual input. Early visual-centric VLAs, such as $\pi_{0.5}$\cite{intelligence2025pi05visionlanguageactionmodelopenworld} and InternVLA-M1 \cite{chen2025internvlam1spatiallyguidedvisionlanguageaction}, address this by leveraging strong vision-language alignment for spatial localization. To further enforce visual attention, recent works explore integrated grounding techniques. ReconVLA~\cite{song2025reconvla} introduces an implicit paradigm that forces a diffusion transformer to reconstruct target gaze regions from visual outputs. Similarly, approaches like InterleaveVLA \cite{fan2025interleave} and 3D-CAVLA~\cite{bhat20253d} attempt to improve scene awareness by interleaving visual tokens with language or incorporating chain-of-thought region detection. However, these integrated methods lack explicit architectural decoupling. By compelling the VLM to jointly process semantic reasoning and specific control trajectories, they remain susceptible to catastrophic forgetting and exhibit limited planner generalization in novel scenarios.

Addressing these coupling issues, explicit hierarchical grounding methods utilize spatial representations as intermediate bridges between perception and action. Systems like DexGraspVLA \cite{zhong2025dexgraspvlavisionlanguageactionframeworkgeneral} and RoboGround~\cite{huang2025roboground} employ visual segmentation masks to isolate target objects and guide downstream policies. While conceptually appealing, generating dense segmentation masks is not a native task for standard VLMs, often requiring external expert models that compromise general visual capabilities. Furthermore, RoboGround~\cite{huang2025roboground} relies on a traditional GR-1~\cite{wu2023unleashing} transformer policy, which struggles to match the continuous control performance of modern Diffusion Transformer(DiT)~\cite{peebles2023scalable} architectures. Similarly, DexGraspVLA~\cite{zhong2025dexgraspvlavisionlanguageactionframeworkgeneral} applies masks to heavily down-sampled global images, diluting the high-fidelity visual details crucial for precise manipulation. These collective shortcomings highlight a critical gap: existing systems fail to effectively bridge the VLM and the action expert using a native, computationally efficient grounded plan. Our proposed HiVLA resolves this by utilizing native VLM bounding boxes to extract high-resolution local crops, which are subsequently fused with global context and explicit skill semantics through a novel cascaded DiT architecture.
\section{Problem Formulation}
\label{sec:problem}
\begin{figure}[tb]
   \centering
   \includegraphics[width=1\textwidth]{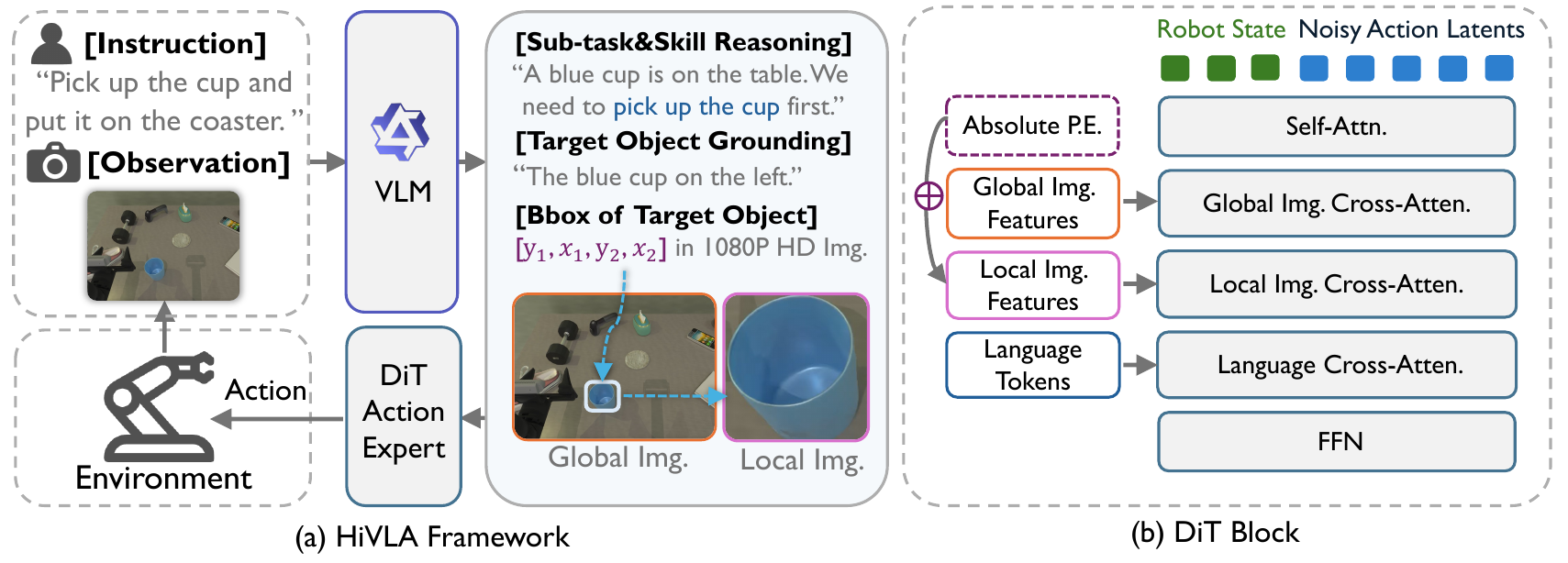}
\caption{\textbf{Pipeline of HiVLA.} (a) Our decoupled framework utilizes a VLM to decompose user instructions into explicit structured plans, yielding a skill-level subtask and a bounding box used to extract a high-resolution target crop. (b) To execute this plan, the DiT action expert employs a cascaded cross-attention block. This design sequentially conditions the noisy action latents on global visual context, position-aware local features, and language tokens, bridging high-level reasoning with low-level control.}
   \label{fig:main}
   \vspace{-5pt}
\end{figure}

We formulate the language-guided robotic manipulation task as a conditional sequence generation problem. The objective is to learn a generalized manipulation system $\pi_{\theta}$ with parameters $\theta$, which maps multi-modal observations and human language instructions to a sequence of executable actions. Formally, at each timestep $t$, the agent receives a set of multi-modal observations $\mathcal{S}_t$. These observations consist of multi-view visual inputs $\mathcal{O}_t = \{I^k_t\}_{k=1}^K$ from $K$ cameras (\textit{e.g.}, wrist and head cameras), providing high-resolution images $I^k_t \in \mathbb{R}^{H \times W \times 3}$ (at a native $1920 \times 1080$ resolution), and the robot's current proprioceptive state $s_t \in \mathbb{R}^{d_s}$ (encoding joint positions and gripper state). Given the history of observations $\mathcal{S}_{0:t} = (\{\mathcal{O}_j\}_{j=0}^t, \{s_j\}_{j=0}^t)$ and the high-level language instruction $L$, the policy $\pi_{\theta}$ aims to generate a sequence of future actions $A_t = \{a_{t+i}\}_{i=0}^{H-1}$ over a prediction horizon $H$. Each action $a_i \in \mathbb{R}^{d_a}$ specifies the target control commands (\textit{e.g.}, joint positions) for the robot's arm and gripper.

The core challenge lies in bridging the semantic gap between the abstract, long-horizon instruction $L$ and the sequence of precise motor commands $A_t$. This difficulty is twofold. First, language instructions demand sophisticated reasoning and task decomposition to translate complex, history-dependent procedures into actionable steps. Second, the visual observations $\mathcal{O}_t$ are inherently challenging, as typical manipulation scenes are cluttered with distractor objects and severe perceptual noise. Ultimately, the system must seamlessly deduce the correct subtask from the instruction, visually ground it onto the target object amidst this background clutter, and translate this grounded intent into robust actions.

\section{Method}
\label{sec:method}

This section introduces HiVLA, a visual-grounded-centric hierarchical manipulation system. We will begin by describing the High-Level VLM Planner (\cref{VLM Planner Agent}), which is responsible for decomposing a given task into actionable subtasks and performing the corresponding visual grounding. Then we will detail the architecture of our DiT-based Action Expert (\cref{sec:action_expert}
), with a focus on its mechanism for effectively conditioning on the high-level plans provided by the VLM. ~\cref{fig:main} illustrates the overall inference process.

\subsection{VLM Planner Agent}
\label{VLM Planner Agent}
The cognitive core of our HiVLA system is a High-Level Planner Agent, implemented with a state-of-the-art Vision-Language Model (VLM). This module serves as the “brain” of the system, responsible for interpreting the high-level language instruction $L$ in the context of the current visual scene $\mathcal{O}_t$ to decide \textit{what} to do next and \textit{where} to do it.

The design of the planner agent is centered around a structured inference process. At each decision step $t$, the agent is provided with the overall goal $L$, the robot's gripper status from state $s_t$, the previous subtask executed, and a visual history comprising the scene before and after the last action. Based on these multi-modal inputs, the VLM reasons about the progress towards $L$ and determines the next logical step. The decomposition strategy is contingent on task complexity; simple instructions may map to a single action, whereas complex, long-horizon tasks (\textit{e.g.}, ``stack three blocks'') are broken down into a sequence of subtasks. Each subtask is typically a pairing of a primitive skill (\textit{e.g.}, `pick', `place') with a single target object. The agent's reasoning culminates in the generation of a structured plan, a JSON object containing the next subtask's description $L_{sub,t}$, the action type, the target object's name, and a normalized bounding box $B_t = [y_{min}, x_{min}, y_{max}, x_{max}] \in ^4$ that localizes the target object in the current scene image.

A key technical advantage of this design is the decoupling of high-level semantic planning from low-level motion generation. By leveraging a pre-trained VLM, the system inherits sophisticated reasoning capabilities, enabling it to handle a wide range of instructions without requiring exhaustive training for every conceivable task. Furthermore, we frame the VLM planner as an intelligent agent that uses tools to execute its intent. The generation of the bounding box $B_t$ is not merely an output; it is a directive that invokes an \textit{Image Crop} tool. This tool uses the normalized coordinates in $B_t$ to extract a high-resolution, object-centric patch $I^{local}_t$ from the original camera observation $I^k_t \in \mathbb{R}^{1080 \times 1920 \times 3}$. The complete, structured plan, including the subtask description $L_{sub,t}$ and the rich visual information from $I^{local}_t$, is then passed as a conditional guidance signal to the DiT Action Expert, which can be conceptualized as the final tool the planner uses to translate its intention into physical action $A_t$.

\subsection{DiT Action Expert}
\label{sec:action_expert}

The DiT Action Expert serves as the manipulation-focused ``hands" of the HiVLA system, responsible for translating the high-level plans formulated by the VLM Planner into precise, low-level motor commands. At its core, this module is a conditional Diffusion Transformer (DiT) designed to model the complex conditional probability distribution $p(A_t | \mathcal{S}_{0:t}, L_{sub,t}, B_t)$. To achieve this, we first detail the continuous-time flow-matching framework, followed by a description of the novel hierarchical transformer architecture that implements it.

\subsubsection{Conditional Flow Matching for Action Generation.}
\label{sec:cfm}
Our objective is to learn a deterministic mapping from a simple noise distribution to the complex data distribution of action sequences, conditioned on the rich context provided by the VLM planner. We employ Conditional Flow Matching (CFM), a powerful generative model that learns to approximate the conditional vector field.

Let $A_t = \{a_{t+i}\}_{i=0}^{H-1}$ be the ground truth action sequence over a horizon $H$, and let the comprehensive conditioning context be denoted by $\mathcal{C}_t = (\mathcal{S}_{0:t}, L_{sub,t}, B_t)$. CFM defines a time-continuous probability path between a sample from a standard Gaussian prior, $\mathbf{z} \sim \mathcal{N}(0, \mathbf{I})$, and the target action sequence $A_t$. We utilize a simple linear interpolation path, defined for a continuous time variable $\tau \in [0, 1]$:
{\small
\begin{equation}
    \mathbf{x}_\tau = \tau A_t + (1-\tau) \mathbf{z}
    \label{eq:cfm_path}
\end{equation}}
At $\tau=0$, the path begins with pure noise ($\mathbf{x}_0 = \mathbf{z}$), and at $\tau=1$, it culminates in the target action sequence ($\mathbf{x}_1 = A_t$).

The neural network, $v_\theta$, which we instantiate as our DiT architecture, is trained to predict the vector field $\mathbf{u} = A_t - \mathbf{z}$ that defines the ``flow'' from noise to data. The training objective is to minimize the L2 distance between the network's prediction and this target vector field, formulated as the following loss function:
\begin{equation}
    \mathcal{L}_{\text{CFM}}(\theta) = \mathbb{E}_{\tau, A_t, \mathbf{z}} \left[ \left\| v_\theta(\mathbf{x}_\tau, \tau, \mathcal{C}_t) - (A_t - \mathbf{z}) \right\|^2 \right]
    \label{eq:cfm_loss}
\end{equation}

During inference, we generate the action sequence by solving the ordinary differential equation (ODE) defined by the learned vector field: $\frac{d\mathbf{x}}{d\tau} = v_\theta(\mathbf{x}_\tau, \tau, \mathcal{C}_t)$. Starting from an initial noise sample $\mathbf{x}_0 \sim \mathcal{N}(0, \mathbf{I})$, we integrate from $\tau=0$ to $\tau=1$. This is approximated using a numerical ODE solver, such as the forward Euler method, over a discrete number of steps:
\begin{equation}
    \mathbf{x}_{\tau+\Delta\tau} = \mathbf{x}_\tau + \Delta\tau \cdot v_\theta(\mathbf{x}_\tau, \tau, \mathcal{C}_t)
    \label{eq:inference_step}
\end{equation}
where $\Delta\tau$ is the step size. This process deterministically transforms the initial noise into a coherent action sequence that is precisely conditioned on context $\mathcal{C}_t$.

\textbf{Hierarchical Transformer Architecture.}
\label{sec:dit_architecture}
The neural network $v_\theta$, which approximates the conditional vector field, is instantiated as a transformer-based architecture. The architecture of our action expert is based on H-RDT, which employs a LLaMA-style transformer backbone featuring RMSNorm for layer normalization and SwiGLU activation functions for enhanced performance. The input to the transformer is a sequence of tokens representing the current proprioceptive state $s_t$ and a noisy future action sequence $\tilde{A}_t$, each projected into the model's hidden dimension $d_{model}$ by dedicated MLP adapters. The diffusion timestep $\tau$ is encoded into a vector embedding and integrated into each transformer block via Adaptive Layer Normalization (AdaLN), which modulates the activations without altering the core feature representations.

The primary innovation of our action expert lies in its hierarchical conditioning mechanism, which is meticulously designed to leverage the rich, multi-faceted plan provided by the VLM planner. Within each transformer block, the model sequentially integrates three distinct forms of guidance through a cascade of cross-attention layers, ensuring a synergistic fusion of global context, local detail, and task-specific instructions.

\paragraph{Global Visual Context.}
The first layer of conditioning provides the model with a comprehensive understanding of the entire scene. The multi-view visual inputs $\mathcal{O}_t$ are processed by a pre-trained vision encoder, a powerful combination of DINOv2 and SigLIP, to produce a set of feature tokens $C^{global} \in \mathbb{R}^{N_{global} \times d_{model}}$. A cross-attention mechanism allows the state-action tokens to attend to these global features. This enables the policy to ground its actions within the broader spatial and semantic context of the environment, performing coarse-grained reasoning about object relationships and the overall workspace layout.

\paragraph{Position-Aware Local Features.}
Following the global context integration, a second, specialized cross-attention layer injects fine-grained, object-centric visual information. This guidance originates from the local image patch $I^{local}_t$, which is cropped from the original high-resolution ($1920 \times 1080$) camera frame using the bounding box $B_t$ supplied by the VLM planner. Cropping from the full-resolution image is critical as it preserves high-fidelity details of the target object that would be lost in down-sampled inputs. After passing $I^{local}_t$ through the same vision encoder to obtain feature tokens $C^{local} \in \mathbb{R}^{N_{local} \times d_{model}}$, we introduce a crucial inductive bias: absolute spatial awareness. For each patch token in $C^{local}$, which corresponds to a specific region in the cropped image, we compute its central coordinate $p \in \mathbb{R}^2$ within the original high-resolution camera frame. This coordinate is then converted into a fixed sinusoidal positional embedding $PE(p) \in \mathbb{R}^{d_{model}}$, inspired by DETR. The final local conditioning signal is formed by element-wise addition:
{\small
\begin{equation}
C^{local-pos} = C^{local} + PE(p)
\end{equation}
}This position-aware feature set provides the model with a detailed, magnified view of the target object while explicitly informing it of the object's precise location in global scene, a critical factor for achieving high-precision manipulation.

\paragraph{Subtask Language Guidance.}
The final conditioning stage aligns the policy with the specific skill required for the current subtask. The subtask description $L_{sub,t}$ from the VLM's plan is encoded into a sequence of language embeddings $C^{lang} \in \mathbb{R}^{N_{lang} \times d_{model}}$. A third cross-attention layer allows the model to attend to these language features, thereby conditioning the generated motion on the precise semantics of required skill (\textit{e.g.}, distinguishing among `pick', `place', or `push').

Finally, after passing through all transformer blocks, the output hidden states corresponding to the action sequence are processed by a final MLP-based Action Decoder. This decoder, also modulated by the timestep embedding, maps the hidden states back to the robot's native action space, producing the denoised action sequence $A_t$. Through this cascaded conditioning strategy, the DiT Action Expert maximally utilizes every component of the VLM's high-level reasoning, effectively grounding abstract plans into robust and precise physical execution.


\section{Experiments}
\label{sec:experiments}

\begin{figure}[tb]
   \centering
   \includegraphics[width=1\textwidth]{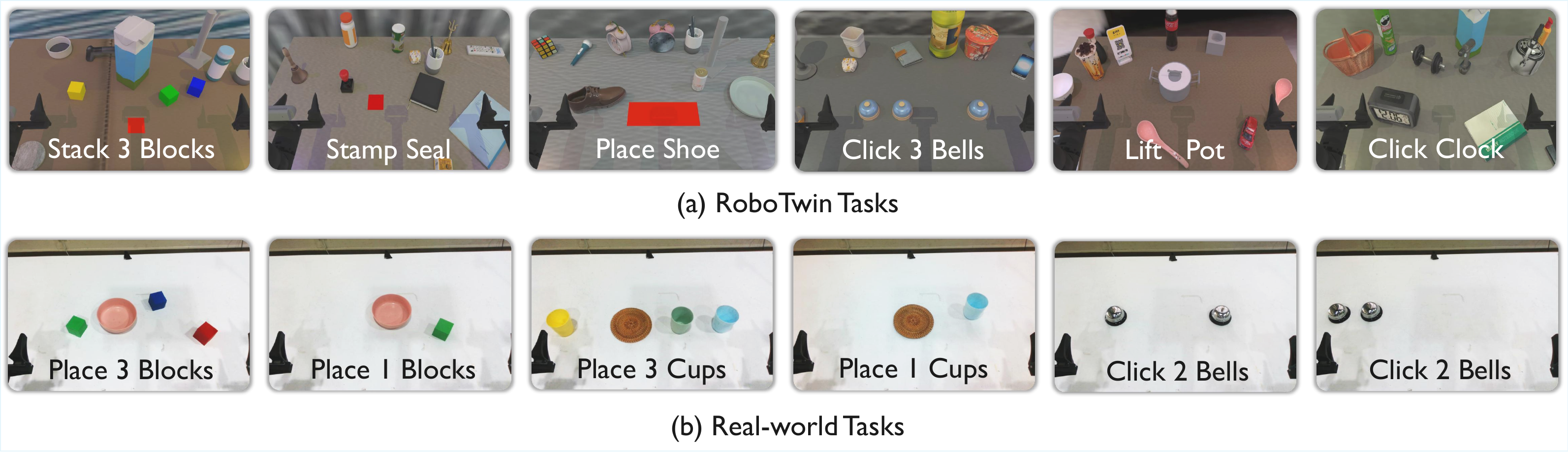}
   \vspace{-15pt}
   \caption{Visualization of RoboTwin tasks and real-world tasks.}
   \label{fig:task_vis}
   \vspace{-10pt}
\end{figure}

\subsection{Experimental Setup}
\label{sec:exp_setup}

To validate the efficacy of the visual-grounded hierarchical design of HiVLA, we conduct extensive experiments in both the RoboTwin2.0~\cite{chen2025robotwin} simulation platform and real-world robotic manipulation settings. We aim to answer the following questions: \textbf{(1)} Does our hierarchical VLA outperform state-of-the-art coupled VLA models? \textbf{(2)} How robust is the control policy to reasoning errors from the high-level planner? \textbf{(3)} How do different visual representations and guidance injection strategies affect the system's performance?

\textbf{Simulation and Robot Configuration.}
We employ RoboTwin2.0~\cite{chen2025robotwin}, a high-fidelity simulation platform specifically designed for robot learning, which facilitates the generation of large-scale datasets and enables reproducible evaluation. To closely emulate the challenges of real-world operation, we utilize the `domain randomization' setting for both data generation and testing. This configuration introduces significant visual diversity and perceptual complexity through randomized backgrounds, cluttered tabletops, variable table heights, and dynamic lighting conditions, thereby posing a rigorous test for visual grounding capabilities. For hardware deployment in both simulation and the real world, we utilize Aloha-Agilex-1.0, a widely-adopted bimanual robot platform featuring a total of 14 Degrees of Freedom (DoF), six per arm plus one for each gripper.

\textbf{Dataset Details.}
We generate a large-scale dataset within the RoboTwin2.0 platform, termed \textbf{HiVLA-HD} (High-Definition). Generated under the `Hard' mode configuration, it comprises 15 manipulation tasks that demand robust visual perception and complex language reasoning. Observations from the head camera are saved at a high resolution of $1920 \times 1080$, while wrist cameras operate at 720p. Leveraging the simulator's capabilities, we obtain precise ground-truth annotations without manual cost: subtask transitions are logged via action planning scripts, and accurate bounding boxes for target objects are derived directly from unique mask IDs. Following rigorous filtering, HiVLA-HD yields approximately 1,000 episodes per task, forming a standardized, high-resolution dataset with fine-grained semantic labels. Crucially, all evaluated models are finetuned on this dataset to ensure a fair comparative analysis.

\begin{table}[tb]
  \centering
  \caption{Main success rates across 9 tasks in the RoboTwin simulator. HiVLA demonstrates superior performance, particularly in long-horizon and visually demanding tasks. Best and second-best results are \textbf{bold} and \underline{underlined}.}
  \label{tab:main_table}
  \setlength{\tabcolsep}{8pt} 
  \resizebox{1\textwidth}{!}{
  \begin{tabular}{@{}l cccccc@{}} 
  \toprule
  
  \textbf{Task} & $\pi_0$~\cite{black2024pi_0} & $\pi_{0.5}$~\cite{intelligence2025pi05visionlanguageactionmodelopenworld} & StarVLA~\cite{starvla2025} & H-RDT~\cite{bi2025h} & Ours w/o Skill & \textbf{Ours} \\
  \midrule
  
  \rowcolor[HTML]{F2F2F2}[0pt][0pt] \multicolumn{7}{@{}l@{}}{\textit{Easy Tasks}} \\
  Click Bell       & 45\% & 65\% & 71\% & 88\% & \textbf{95\%} & \underline{94\%} \\
  Click Clock      & 53\% & 66\% & 83\% & 93\% & \textbf{97\%} & \textbf{97\%} \\
  Press Stapler    & 60\% & 69\% & 63\% & 89\% & \textbf{98\%} & \underline{97\%} \\
  Lift Pot         & 59\% & 21\% & 18\% & 92\% & \textbf{96\%} & \textbf{96\%} \\
  \textbf{\textit{Average}} & 54.3\% & 55.3\% & 58.8\% & 90.5\% & \textbf{96.5\%} & \underline{96.0\%} \\
  \midrule
  
  \rowcolor[HTML]{F2F2F2}[0pt][0pt] \multicolumn{7}{@{}l@{}}{\textit{Hard Tasks}} \\
  Place Shoe       & 75\% & 68\% & 61\% & 88\% & \underline{94\%} & \textbf{95\%} \\
  Move Stapler     & 15\% & 17\% & 15\% & 34\% & \underline{42\%} & \textbf{60\%} \\
  Stamp Seal       & 61\% & 42\% & 25\% & 43\% & \underline{68\%} & \textbf{76\%} \\
  Stack 3 Blocks   & 1\%  & 1\%  & 16\% & 20\% & \underline{26\%} & \textbf{37\%} \\
  Click 3 Bells    & 41\% & 54\% & 66\% & 88\% & \underline{92\%} & \textbf{98\%} \\
  \textbf{\textit{Average}} & 38.6\% & 36.4\% & 36.6\% & 54.6\% & \underline{64.4\%} & \textbf{73.2\%} \\
  \midrule
  
  \textbf{\textit{Total Average}} & 45.6\% & 44.8\% & 46.4\% & 70.6\% & \underline{78.7\%} & \textbf{83.3\%} \\
  \bottomrule
  \end{tabular}}
  \vspace{-5pt}
\end{table}

\textbf{Baseline Selection.}
To comprehensively evaluate our approach, we benchmark HiVLA against four state-of-the-art (SOTA) models: $\pi_0$~\cite{black2024pi_0}, its advanced variant $\pi_{0.5}$~\cite{intelligence2025pi05visionlanguageactionmodelopenworld}, StarVLA~\cite{starvla2025}, and H-RDT~\cite{bi2025h}. $\pi_0$ and $\pi_{0.5}$ represent SOTA dual-system VLAs that accomplish perception and reasoning through joint training and parallel inference. StarVLA provides a comprehensive suite of mainstream VLA architectures built upon the Qwen-VL~\cite{bai2025qwen3} backbone. Specifically, we evaluate its Qwen-GR00T variant (whose performance officially matches GR00T-N1.5~\cite{bjorck2025gr00t}), equipped with the exact same Qwen3-VL backbone as our framework, to ensure a strictly fair comparison of the architectural paradigms. H-RDT serves as a critical baseline and an implicit ablation of our visual-grounding mechanism, as it relies entirely on global image features for policy generation. 

We note that open-source, visual-grounded hierarchical systems designed for general manipulation remain scarce, which informs our baseline choices. For instance, DexGraspVLA~\cite{zhong2025dexgraspvlavisionlanguageactionframeworkgeneral} is strictly restricted to multi-object grasping and lacks cross-task generalization, precluding a feasible comparison. Proprietary systems like Gemini Robotics~\cite{gemini_robotics_2025} and HiRobot~\cite{shi2025hirobotopenendedinstruction} are closed-source. Furthermore, InterleaveVLA~\cite{fan2025interleave} focuses primarily on object-centric policies rather than offering a complete hierarchical system.

\textbf{System Latency Analysis.}
Our decoupled architecture resolves the inherent frequency mismatch between slow vision-language reasoning and high-speed motor control via asynchronous inference. While the unoptimized VLM Planner requires 1.9s per reasoning step, leaving ample room for software acceleration, the DiT Action Policy efficiently infers a 16-step action chunk in merely 0.162s. By executing the semantic planner in parallel with the fast control policy and maintaining temporal consistency through mechanisms like real-time tracking, the system effectively bridges this latency gap to achieve an $8\text{Hz}$ control frequency, demonstrating strong practicality for real-world deployment.

\subsection{Evaluation in RoboTwin Platform}
\label{sec:sim_eval}

\textbf{Evaluation Tasks and Protocols.}
For a comprehensive assessment, we benchmark all models across a suite of 9 tasks, categorized into four \textit{Easy Tasks} and five \textit{Hard Tasks}. Easy tasks typically require a single skill and evaluate precise visual perception (\textit{e.g.}, grasping small objects like a bell or a stapler). Hard tasks involve sequences of multiple skills or demand advanced spatial and semantic reasoning. For instance, `Stack 3 Blocks' requires the model to sequentially infer the correct colored block to manipulate based on a specific visual order, while `Click 3 Bells' presents three identical bells, forcing the model to rely strictly on spatial language reasoning (`left', `center', `right') to disambiguate the target. Visualizations of these selected tasks are provided in~\cref{fig:task_vis}, with a comprehensive list of textual instructions and visual states detailed in the Appendix. For each task, we conduct 100 independent trials under unseen environment configurations, reporting the average success rate over the last three saved checkpoints to ensure statistical stability.

\textbf{VLM Planner Training and Validation.}
To validate the high-level reasoning capabilities of our VLM Planner, we curated a specialized 210K-instance dialogue dataset derived from HiVLA-HD. Fine-tuning the Qwen3-VL~\cite{bai2025qwen3} 8B model on this domain-specific data yields highly robust semantic planning, achieving a bounding box grounding accuracy (mIoU) of 90.37\% and a strict exact-match sub-task prediction of 98.57\% (with comprehensive evaluations detailed in the Appendix). Crucially, our decoupled architecture makes the VLM planner readily replaceable. This flexibility allows the system to either undergo lightweight fine-tuning for specific operational domains, or directly integrate off-the-shelf VLMs for zero-shot deployment. While we adopt fine-tuned 8B model for all subsequent evaluations to maximize task performance, this extensible design easily accommodates future advancements in more powerful VLM agents.

\textbf{Action Policy Training and Main Results.}
To ensure a fair comparison, all baselines are all fine-tuned on the HiVLA-HD dataset for 150K steps using two H200 GPUs (batch size 64). Notably, HiVLA's DiT is initialized from H-RDT weights pre-trained on the EgoDex dataset, where we directly copy the weights from the global image cross-attention layer to initialize our novel local image cross-attention layer. 

As presented in~\cref{tab:main_table}, HiVLA achieves an unparalleled total average success rate of 83.3\%, significantly outperforming SOTA coupled VLAs ($\pi_0$, $\pi_{0.5}$, StarVLA) and the purely global-vision-based H-RDT. For \textit{Easy Tasks}, HiVLA's use of high-resolution, object-centric crops explicitly preserves critical visual features, securing a 96.0\% average success rate and showing distinct advantages on small targets. For \textit{Hard Tasks}, the performance gap widens dramatically. Coupled VLAs struggle to maintain spatial temporal consistency over long horizons (averaging $<40\%$). In contrast, HiVLA leverages the high-level planner to comprehend task progression and localize targets, resulting in a commanding 73.2\% average success rate—a remarkable 18.6\% absolute improvement over H-RDT.

\begin{table}[tb]
  \caption{Robustness evaluation of the Action Expert against guidance perturbations. The policy is highly resilient to Bbox noise but strictly adheres to language instructions.}
  \label{tab:noise_robustness}
  \centering
  \small
  \setlength{\tabcolsep}{8pt}
  \resizebox{0.85\textwidth}{!}{
  \begin{tabular}{@{}lcccccc@{}}
  \toprule
  \multirow{2}{*}{\textbf{Corrupted Modality}} & \multicolumn{6}{c}{\textbf{Error Injection Rate}} \\
  \cmidrule(l){2-7}
   & \textbf{0\%} & \textbf{20\%} & \textbf{40\%} & \textbf{60\%} & \textbf{80\%} & \textbf{100\%} \\
  \midrule
  Noise on BBox & 83.3\% & 78.5\% & 74.0\% & 70.8\% & 59.8\% & 57.0\% \\
  Noise on Task (Lang) & 83.3\% & 69.3\% & 49.3\% & 36.0\% & 24.3\% & 12.0\% \\
  Noise on Both & 83.3\% & 62.0\% & 42.5\% & 33.3\% & 23.0\% & 17.3\% \\
  \bottomrule
  \end{tabular}}
  \vspace{-6pt}
\end{table}

\textbf{Skill Decomposition and Error Correction.}
Our ablation variant, Ours (w/o Skill), further isolates the impact of granular language instructions. While performance is comparable on Easy tasks (where the global instruction intrinsically matches a single skill), replacing specific sub-task skills with the global instruction on Hard tasks causes an 8.8\% performance drop. This confirms that decomposed, ``one-to-one" language conditions drastically reduce the cognitive load on the diffusion policy, allowing it to focus purely on local geometry and execution. Furthermore, we observed a compelling \textit{emergent error-correction} property: if the DiT policy fails a grasp (a ``phantom execution"), the VLM Planner acts as an independent semantic supervisor. Recognizing the sub-task as incomplete, it seamlessly re-issues the visual-language command, enabling the system to re-attempt the skill—a resilience unattainable in standard coupled VLAs.

\textbf{Robustness to Planner Errors.}
A pervasive critique of hierarchical systems is their susceptibility to compounding errors, where a planner's mistake irreversibly crashes the downstream policy. To address this, we subjected our Action Expert to rigorous perturbation testing (\cref{tab:noise_robustness}) by injecting calibrated noise into the inferred bounding boxes and language instructions. The results reveal a highly desirable decoupling: the policy demonstrates strong resilience to spatial noise. Even with 100\% shifting in the target bounding box, it retains a 57.0\% success rate, effectively leveraging the auxiliary global image features to self-correct and localize the true target. Conversely, injecting noise into the language instructions causes a proportional degradation in performance, precisely matching the error injection rate. This confirms the policy's strict semantic compliance, proving that our architecture successfully balances robust visual adaptability with absolute adherence to language commands.

\subsection{Evaluation in the Real World}
\label{sec:real_eval}

To further validate real-world applicability and robustness of our visual-grounded hierarchical design, we deployed HiVLA in a physical robotic environment. 

\textbf{Real-World Task Setup and Protocol.}
As visualized in~\cref{fig:task_vis}, our experiments span 7 object categories and encompass 16 distinct sub-type scenarios. Rather than evaluating on conventional easy tasks, we specifically designed these scenarios to stress-test strong cross-environmental generalization and precise instruction following. By utilizing complex combinations of objects with varying colors and spatial arrangements, the tasks range from manipulating a single primitive to selecting a specific target (\eg, a red block or a green cup) from dense, multi-object clutter.

\begin{table}[tb]
  \caption{Real-world success rates. HiVLA excels in multi-object cluttered scenarios requiring semantic grounding, where baseline models struggle.}
  \vspace{-5pt}
  \label{tab:task_success_rate}
  \centering
  \setlength{\tabcolsep}{8pt} 
  \resizebox{0.86\textwidth}{!}{
  \begin{tabular}{@{}lcccccc@{}} 
    \toprule
    \multirow{2}{*}{\textbf{Method}} & \multicolumn{2}{c}{\textbf{Click Bell}} & \multicolumn{2}{c}{\textbf{Pick \& Place Cup}} & \multicolumn{2}{c}{\textbf{Pick \& Place Block}} \\
    \cmidrule(lr){2-3} \cmidrule(lr){4-5} \cmidrule(lr){6-7}
    & 1 Bell & 2 Bells & 1 Cup & 3 Cups & 1 Block & 3 Blocks \\
    \midrule
    H-RDT & 8/30 & 9/30 & 4/30 & 0/30 & 9/30 & 0/30 \\
    \textbf{Ours} & 13/30 & 17/30 & 21/30 & 6/30 & 20/30 & 7/30 \\
  \bottomrule
  \end{tabular}}
  \vspace{-5pt}
\end{table}

For model training, we collected a dataset of 360 teleoperated episodes, automatically annotated with precise bounding boxes via GroundingDINO~\cite{liu2024grounding} and SAM2~\cite{ravi2024sam}. Both H-RDT and HiVLA were initialized from their simulation-trained checkpoints to leverage structural priors, and subsequently fine-tuned for 80K steps on real-world data. During the evaluation phase, each task was attempted for 30 trials, with object positions randomized to prevent rote memorization and rigorously assess policy robustness.

\textbf{Success Rate Analysis.}
The real-world success rates are detailed in~\cref{tab:task_success_rate}. It is crucial to note that our evaluation suite intentionally targets hard, strong-generalization scenarios requiring rigorous semantic reasoning. Consequently, the baseline H-RDT exhibits performance degradation. While it performs adequately in isolated single-object scenarios, its success rate collapses to nearly zero in multi-object clutter (\textit{e.g.}, `3 Cups', `3 Blocks'). Relying solely on global visual features, H-RDT lacks the fine-grained grounding required to disambiguate identical shapes using color attributes or spatial commands. In contrast, HiVLA's hierarchical decoupling relieves the DiT Action Expert of the global reasoning burden, allowing it to efficiently map the sparse, limited real-world data to precise local visual-language conditions. As a result, HiVLA effectively navigates complex, cluttered scenes, executing sub-skills with remarkable accuracy and generalizing robustly across demanding physical tasks.

\subsection{Ablation Study}
\label{sec:ablation}

We conduct ablation studies on the DiT Action Expert's architecture to analyze the efficacy of visual representations and cross-attention guidance strategies.

\textbf{Guidance Injection Strategy.}
The order in which conditions are injected into the DiT via cross-attention heavily impacts policy learning. We expanded our ablation to isolate guidance contributions (\cref{tab:ablation_combined}). Relying solely on Local or Global visual features yields suboptimal results ($\sim$70\% average). Combining both captures the broader environment while explicitly grounding the target. More importantly, our analysis of the cross-attention ordering confirms that the ``Coarse-to-Fine'' injection strategy (\textit{Global Context $\to$ Local Crop $\to$ Language Skill}) allows the DiT to progressively narrow its attention from the entire scene to the specific object, and finally to the semantic action, yielding the optimal $83.3\%$ average success rate.

\begin{table}[tb]
  \centering
  \caption{Ablation study on guidance injection strategies and visual-grounding components. Best results are \textbf{bold}.}
           \vspace{-5pt}
  \label{tab:ablation_combined}

  \footnotesize
  \setlength{\tabcolsep}{3pt}

  \resizebox{0.9\textwidth}{!}{
  \begin{tabular}{@{} l ccccccccc @{\hspace{1.5ex}} c @{}}
  \toprule
  \raisebox{1.5em}{\textbf{Method}}
    & \rotatebox{90}{\shortstack[c]{Click \\ Bell}}
    & \rotatebox{90}{\shortstack[c]{Click \\ Clock}}
    & \rotatebox{90}{\shortstack[c]{Press \\ Stapler}}
    & \rotatebox{90}{\shortstack[c]{Lift \\ Pot}}
    & \rotatebox{90}{\shortstack[c]{Place \\ Shoe}}
    & \rotatebox{90}{\shortstack[c]{Move \\ Stapler}}
    & \rotatebox{90}{\shortstack[c]{Stamp \\ Seal}}
    & \rotatebox{90}{\shortstack[c]{Stack \\ Blocks}}
    & \rotatebox{90}{\shortstack[c]{Click \\ 3 Bells}}
    & \raisebox{1.5em}{\textbf{Avg}} \\
  \midrule

  \rowcolor[HTML]{F2F2F2}[0pt][0pt]
  \multicolumn{11}{@{}l@{}}{\textit{(A) Guidance Injection Strategy}} \\

  Local$\to$Text
    & 93\% & 95\% & 96\% & 89\% & 76\% & 30\% & 60\% & 12\% & 84\% & 70.4\% \\
  Global$\to$Text
    & 88\% & 93\% & 89\% & 92\% & 88\% & 34\% & 43\% & 20\% & 88\% & 70.6\% \\
  \midrule
  Local$\to$Text$\to$Global
    & \textbf{97\%} & 92\% & 97\% & \textbf{98\%} & 90\% & 49\% & 69\% & \textbf{43\%} & 88\% & 80.1\% \\
  Global$\to$Text$\to$Local
    & 90\% & 88\% & 95\% & \textbf{98\%} & 94\% & 48\% & 65\% & 39\% & 89\% & 78.3\% \\
  Local$\to$Global$\to$Text
    & 89\% & \textbf{97\%} & \textbf{99\%} & \textbf{98\%} & 92\% & 41\% & 59\% & 17\% & 79\% & 74.1\% \\
  \midrule
  \textbf{Global$\to$Local$\to$Text}
    & 94\% & \textbf{97\%} & 97\% & 96\% & \textbf{95\%} & \textbf{60\%} & \textbf{76\%} & 37\% & \textbf{98\%} & \textbf{83.3\%} \\

  \midrule

  \rowcolor[HTML]{F2F2F2}[0pt][0pt]
  \multicolumn{11}{@{}l@{}}{\textit{(B) Visual-Grounding Components}} \\

  w/o HD Crop
    & {85\%} & {92\%} & {94\%} & 79\% & {94\%} & 57\% & 63\% & {31\%} & 82\% & 75.2\% \\
  w/o Abs.\ PE
    & {90\%} & {92\%} & {95\%} & 93\% & {94\%} & 52\% & 62\% & {33\%} & 80\% & 76.8\% \\
  \midrule
  \textbf{Ours (Full)}
    & \textbf{94\%} & \textbf{97\%} & \textbf{97\%} & \textbf{96\%} & \textbf{95\%}
    & \textbf{60\%} & \textbf{76\%} & \textbf{37\%} & \textbf{98\%} & \textbf{83.3\%} \\

  \bottomrule
  \end{tabular}}
  \vspace{-5pt}
\end{table}

\textbf{Visual-Grounding Components.} 
We investigate two variants: \textbf{(1) Low-Res Crop}: Cropping from a down-sampled $640 \times 360$ image rather than the 1080p source. \textbf{(2) w/o Absolute PE}: Removing absolute sinusoidal positional encoding for the cropped image tokens. As shown in \cref{tab:ablation_combined}, low-resolution crops significantly degrade performance on tasks involving fine-grained structures (\textit{e.g.}, grasping the thin handle in `Lift Pot'). Furthermore, without absolute spatial PE, the model fails to disambiguate identical objects (\textit{e.g.}, `Click 3 Bells'), proving that explicit spatial guidance is indispensable.

\section{Conclusion}
\label{sec:conclusion}

In this work, we presented HiVLA, a hierarchical visual-grounded-centric manipulation system that effectively resolves the fundamental trade-off in end-to-end VLA models between preserving VLM reasoning capabilities and achieving precise low-level control. By decoupling high-level planning from action generation, our framework employs a VLM planner for task decomposition and visual grounding, while a novel DiT action expert leverages this grounded plan through a cascaded cross-attention mechanism that integrates global context, position-aware local features, and subtask guidance. Extensive experiments in both simulation and real-world settings demonstrate that HiVLA significantly outperforms state-of-the-art baselines, achieving an 12.7\% improvement over H-RDT and 37.7\% over $\pi_0$ in simulation, with particular strength in long-horizon skill composition and fine-grained manipulation of small objects in cluttered environments. Beyond performance gains, HiVLA's modular architecture enables independent scaling of each component and provides interpretability through explicit intermediate plans, establishing a robust and scalable foundation for complex robotic manipulation systems.



%
%
\bibliographystyle{splncs04}
\bibliography{main}
\clearpage
\setcounter{section}{0}
\begin{center}
    {\Large\bfseries Supplementary Material}
\end{center}
\section{DiT Model Details}
\label{sec:details}

\subsubsection{Implementation Details}
We implemented our model using the PyTorch framework, leveraging the HuggingFace Accelerate library for distributed training. The model was trained on a cluster equipped with 2 NVIDIA H200 GPUs. We utilized the AdamW~\cite{loshchilov2017decoupled} optimizer with a weight decay of $1 \times 10^{-2}$ and a gradient clipping threshold of 1.0 to ensure training stability. The learning rate followed a constant schedule with a linear warmup phase of 500 steps, peaking at $1 \times 10^{-4}$. To optimize memory usage and computational throughput without compromising performance, we employed BFloat16 (BF16) mixed-precision training. The global batch size was set to 64 (32 per GPU). The model was trained for 150k steps. ~\cref{tab:arch_details} presents the detailed hyperparameters and architecture specifications for the DiT Action Expert.

\subsubsection{Architecture Specifications}
Our architecture follows a high-capacity Transformer design. The core backbone consists of 16 layers with a hidden dimension of 2,176. We employed Grouped Query Attention (GQA) to balance computational efficiency and performance, utilizing 16 attention heads and 8 key-value heads. For the feed-forward networks (FFN), we adopted the SwiGLU activation function, following the architectural patterns of LLaMA~\cite{touvron2023llama}. Layer Normalization (LayerNorm) with an epsilon of $1 \times 10^{-5}$ was applied before the attention and FFN blocks (Pre-LN).

\begin{table}[tb]
  \centering
  \caption{\textbf{Hyperparameters and architecture specifications.} Detailed configuration of the HiVLA DiT Action Expert and training settings.}
  \label{tab:arch_details}
  \setlength{\tabcolsep}{8pt}
  \resizebox{0.7\linewidth}{!}{
  \begin{tabular}{@{}lc@{}}
    \toprule
    \textbf{Hyperparameter} & \textbf{Value} \\
    \midrule

    \rowcolor[HTML]{F2F2F2}[0pt][0pt] \multicolumn{2}{@{}l@{}}{\textit{Transformer Backbone}} \\
    Hidden size & 2176 \\
    Layers & 16 \\
    Attention heads & 16 \\
    Key-value heads (GQA) & 8 \\
    Activation function & SwiGLU \\
    Normalization & LayerNorm (\(\epsilon = 1\mathrm{e}{-5}\)) \\
    Action chunk size (horizon) & 16 \\
    \midrule

    \rowcolor[HTML]{F2F2F2}[0pt][0pt] \multicolumn{2}{@{}l@{}}{\textit{Encoders \& adapters}} \\
    Vision backbone & DINO-SigLIP (frozen) \\
    Vision/text adapter & MLP (2-layer, SiLU) \\
    State/action adapter & MLP (3-layer, SiLU) \\
    State/action dimension & 14 \\
    Image resolution & \(384 \times 384\) \\
    Patch size & \(14 \times 14\) \\
    \midrule

    \rowcolor[HTML]{F2F2F2}[0pt][0pt] \multicolumn{2}{@{}l@{}}{\textit{Optimization \& training}} \\
    Optimizer & AdamW \\
    Learning rate & \(1 \times 10^{-4}\) \\
    LR schedule & Constant with warmup \\
    Warmup steps & 500 \\
    Weight decay & \(1 \times 10^{-2}\) \\
    Gradient clipping & 1.0 \\
    Global batch size & 64 \\
    Mixed precision & bfloat16 \\
    Training steps & 150,000 \\
    \bottomrule
  \end{tabular}}
\end{table}

\subsubsection{Input Conditioning}
The vision backbone (DINOv2~\cite{oquab2023dinov2} + SigLIP~\cite{tschannen2025siglip}) was kept frozen during training to leverage robust pre-trained representations. We utilized distinct Multi-Layer Perceptron (MLP) projectors to map different modalities into the transformer's latent space. Specifically, a 2-layer MLP with SiLU activation was used for visual and language embeddings, while a deeper 3-layer MLP was employed for state and action embeddings to capture complex kinematic dynamics.

\section{VLM Planner Agent Analysis}
\label{sec:vlm_planner_analysis}

In this section, we provide a comprehensive analysis of the High-Level VLM Planner Agent. We detail its experimental setup, evaluate the impact of fine-tuning across different model scales, and ablate key design choices such as visual history injection.

\subsubsection{Experimental Setup and Metrics.}
We employ Qwen3-VL~\cite{bai2025qwen3} as our core VLM Planner Agent. It offers formidable perception and reasoning capabilities while maintaining deployment flexibility. To independently assess its planning proficiency, we curated a dataset of 210K dialogue instances derived from HiVLA-HD. This dataset is split into an 80/20 ratio for training and testing. 
For fine-tuning, we trained the models on two NVIDIA H200 GPUs. We used a batch size of 4 and a learning rate of 1e-5, training for 3 epochs. 
During evaluation, we measure visual grounding using the mean Intersection over Union (mIoU) of the predicted bounding boxes. For subtask prediction, we employ a strict exact-match criterion. The model must correctly predict both the required skill and the target object name to score a success.

\begin{table}[htbp]
\centering
\caption{\textbf{Comprehensive Evaluation of the VLM Planner Agent.} We compare different model scales and architectures under zero-shot and fine-tuned settings. Fine-tuning drastically improves domain-specific performance, while historical visual context is critical for optimal accuracy.}
\label{tab:planner_comprehensive}
\resizebox{1\linewidth}{!}{
\begin{tabular}{llcc} 
\toprule
\textbf{Model} & \textbf{Setting} & \textbf{Grounding (mIoU $\uparrow$)} & \textbf{Sub-task Acc. ($\uparrow$)} \\
\midrule
\rowcolor{gray!10} \multicolumn{4}{c}{\textit{Zero-Shot Evaluation}} \\
Qwen3-VL-4B         & Zero-shot & 28.03 & 45.51 \\
Qwen3-VL-8B         & Zero-shot & 12.68 & 35.71 \\
Qwen3-VL-32B        & Zero-shot & 20.17 & 39.85 \\
Qwen3-VL-30B-A3B    & Zero-shot & 32.46 & 41.41 \\
GPT-4o              & Zero-shot &  3.45 & 42.85 \\
\midrule
\rowcolor{gray!10} \multicolumn{4}{c}{\textit{Fine-Tuned Evaluation}} \\
Qwen3-VL-4B         & Fine-tuned & 92.21 & 97.92 \\
Qwen3-VL-8B         & Fine-tuned (w/o history) & 89.63 & 95.24 \\
\textbf{Qwen3-VL-8B}& \textbf{Fine-tuned (Ours)} & \textbf{90.37} & \textbf{98.57} \\
\bottomrule
\end{tabular}
}
\end{table}

\subsubsection{Fine-Tuning vs. Zero-Shot Capabilities.}
We evaluate multiple models under both zero-shot and fine-tuned settings. The comprehensive results are reported in \cref{tab:planner_comprehensive}. 
While baseline VLMs possess competent zero-shot reasoning, their out-of-the-box performance is insufficient for precise, long-horizon manipulation. Scaling up the model parameters (e.g., from 8B to 32B or using MoE architectures like 30B-A3B) steadily improves zero-shot subtask accuracy and grounding. Even proprietary state-of-the-art models like GPT-4o achieve competitive zero-shot subtask accuracy (42.85\%), though their native spatial grounding remains weak (3.45\% mIoU).

Crucially, lightweight fine-tuning on domain-specific visual-language data triggers a massive performance boost. The fine-tuned Qwen3-VL 8B model achieves an exceptional 90.37\% mIoU and 98.57\% subtask accuracy. This underscores a key advantage of our hierarchical design. It preserves the generalizable priors of pretrained VLMs, yet allows for specialized, highly effective enhancements through scalable fine-tuning. To best validate our DiT Action Expert, we utilize this fine-tuned 8B model for all main paper evaluations.

\subsubsection{The Necessity of Visual History.}
Manipulation tasks are inherently sequential. A robust planner must understand what has already been accomplished. To validate this, we ablate the visual history input. As shown in \cref{tab:planner_comprehensive}, removing historical frames during fine-tuning (``w/o history'') leads to a clear performance drop. Subtask accuracy falls from 98.57\% to 95.24\%. This confirms that historical observations are essential. They provide the necessary temporal context for accurate task progression and target disambiguation.

\subsubsection{Extensibility and Future Scaling.}
Our decoupled architecture makes the VLM planner readily replaceable. Advanced foundation models can serve as direct, plug-and-play replacements for the planner module. While domain-specific fine-tuning remains the most optimal deployment strategy today, the steady improvement in zero-shot capabilities of larger models (e.g., Qwen3-VL-32B, GPT-4o) highlights the strong future potential of our system. As VLM agents continue to evolve, HiVLA will seamlessly inherit their enhanced cognitive limits.

\subsubsection{Prompt Design.}
We provide the detailed prompt structure used for the VLM Planner Agent in \cref{tab:vlm_prompt}. The prompt is designed to enforce strict JSON output formatting while providing the agent with context-aware visual and state inputs.

\begin{table*}[t]
  \centering
  \caption{\textbf{System prompt for the VLM planner agent.} The agent receives historical and current observations together with state information, and generates a structured subtask plan.}
  \label{tab:vlm_prompt}
  \small
  \setlength{\fboxsep}{6pt}
  \noindent\fbox{%
    \begin{minipage}{0.965\textwidth}
      \textbf{\textsf{Role}}\\
      You are the central control unit for a robotic arm. Your goal is to analyze visual and state information to decide the next action needed to complete a high-level task.

      \vspace{0.4em}
      \hrule
      \vspace{0.4em}

      \textbf{\textsf{Provided Information}}\\[-0.2em]
      You are given two images in order:
      \begin{enumerate}
        \setlength{\topsep}{2pt}
        \setlength{\itemsep}{1pt}
        \setlength{\parsep}{0pt}
        \setlength{\partopsep}{0pt}
        \item \textbf{Previous Scene Image}: The scene after the last action was executed (corresponding to the first \texttt{<image>}).
        \item \textbf{Current Scene Image}: The live scene right now (corresponding to the second \texttt{<image>}).
      \end{enumerate}

      \vspace{0.2em}
      \hrule
      \vspace{0.4em}

      \textbf{\textsf{Current State Inputs}}\\[-0.35em]
      \begin{itemize}
        \setlength{\topsep}{2pt}
        \setlength{\itemsep}{1pt}
        \setlength{\parsep}{0pt}
        \setlength{\partopsep}{0pt}
        \item \textbf{Overall Goal}: \texttt{\{task\_instruction\}}
        \item \textbf{Previous Subtask Commanded}: \texttt{\{previous\_subtask\}}
        \item \textbf{Current Gripper State}: \texttt{\{gripper\_state\_str\}}
      \end{itemize}

      \vspace{0.2em}
      \hrule
      \vspace{0.4em}

      \textbf{\textsf{Your Task}}\\
      Based \textit{only} on the \textbf{Current State Inputs} and the
      \textbf{two provided images}, you must generate a JSON object
      describing the \textit{next} action to perform.

      Your response \textbf{must be a single JSON object} with no extra
      text or explanations. The JSON object must contain exactly the
      following four keys:

      \vspace{0.4em}
      \fbox{%
        \begin{minipage}{0.94\linewidth}
\ttfamily\scriptsize
\{\\
\hspace*{1em}"next\_subtask\_description": "A clear description of the next
subtask you are planning.",\\[0.2em]
\hspace*{1em}"action\_type": "pick or place",\\[0.2em]
\hspace*{1em}"target\_object": "The specific object involved in the action.
For pick, this is the object to grasp. For place, this is the object that the
robot should place the grasped object onto.",\\[0.2em]
\hspace*{1em}"bbox": "[ymin, xmin, ymax, xmax], a normalized bounding box with
coordinates in [0,1000] for the target\_object in the Current Scene Image."\\
\}
        \end{minipage}
      }
    \end{minipage}
  }
\end{table*}

\section{Task Visualization}
\begin{figure*}[!t] 
   \centering
   \includegraphics[width=1\textwidth]{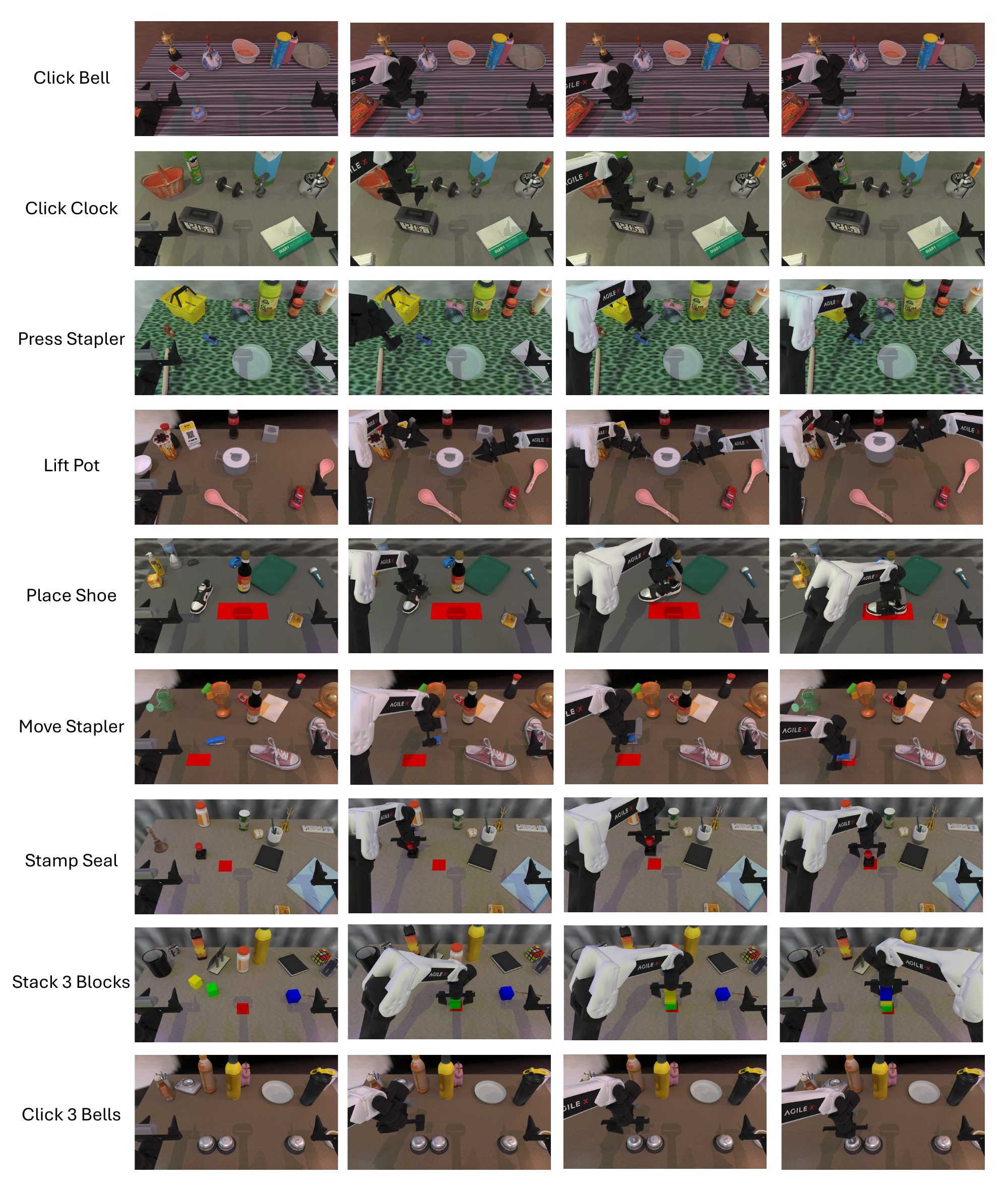}
   \caption{Visualization of RoboTwin tasks.}
   \label{fig:robo_supp}
\end{figure*}

\begin{figure*}[!t]
   \centering
   \includegraphics[width=1\textwidth]{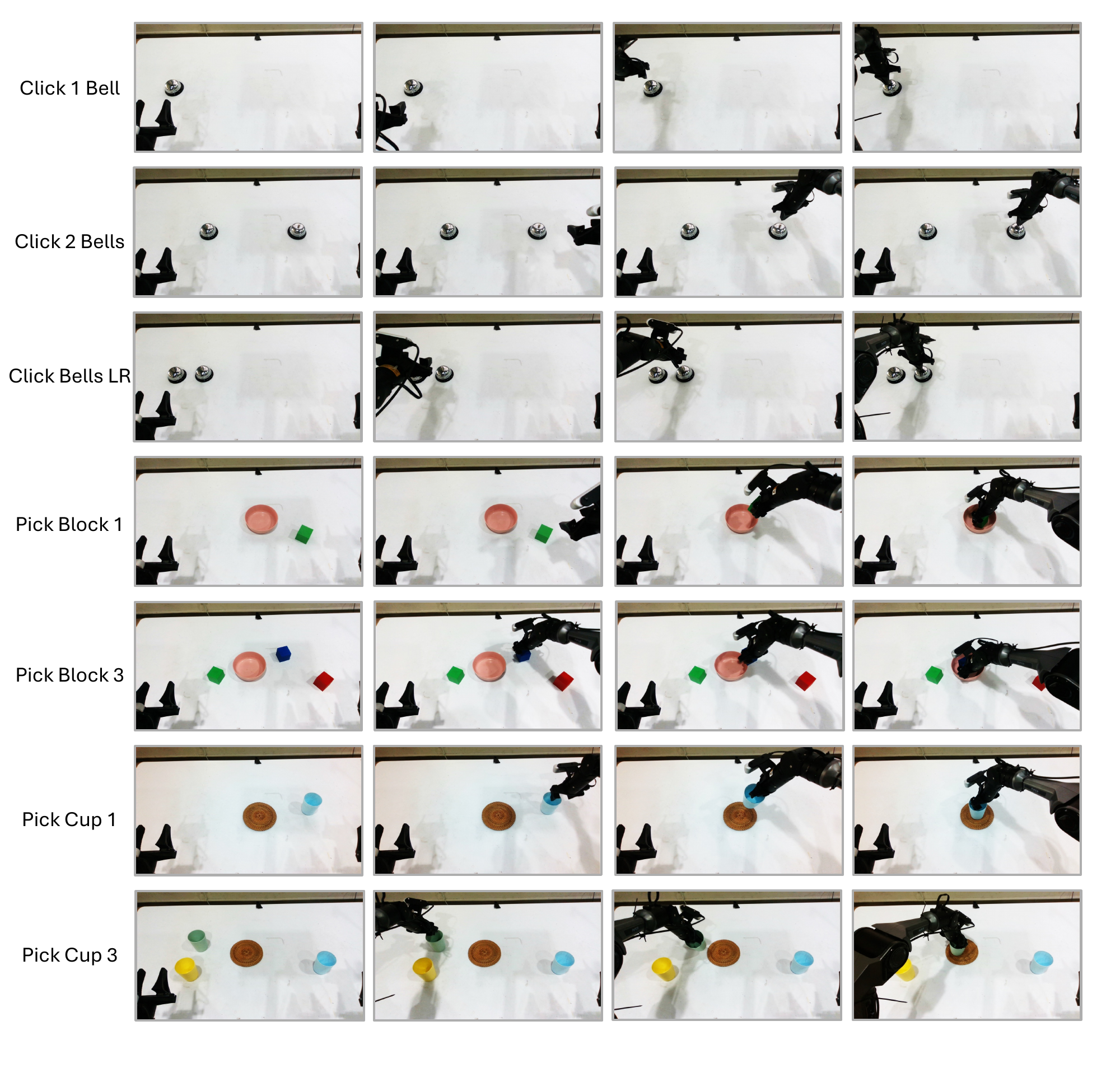}
   \caption{Visualization of real-world tasks.}
   \label{fig:real_supp}
\end{figure*}

We present comprehensive visualizations of the experimental tasks conducted in both the RoboTwin simulation environment and real-world scenarios. The specific natural language instructions corresponding to each task are detailed in \cref{tab:task_instructions}. Visual demonstrations of the execution sequences in the RoboTwin simulation are illustrated in \cref{fig:robo_supp}, while the corresponding real-world execution processes are depicted in \cref{fig:real_supp}.

\begin{table*}[t]
\centering
\caption{\textbf{List of Task Instructions.} The specific natural language instructions corresponding to each task in the RoboTwin simulation and real-world experiments.}
\label{tab:task_instructions}
\small 
\renewcommand{\arraystretch}{1.2} 
\begin{tabular}{@{}l p{0.75\linewidth}@{}} 
\toprule
\textbf{Task Name} & \textbf{Instruction} \\
\midrule
\multicolumn{2}{@{}l}{\textit{\textbf{RoboTwin Simulation Tasks}}} \\ 
\midrule
Click Bell & Click the bell on the table. \\
Click Alarm Clock & Click the alarm clock's center of the top side button on the table. \\
Press Stapler & Use one arm to press the stapler. \\
Lift Pot & Use BOTH arms to lift the pot. \\
Place Shoe & Grab the shoe from the table and place it on the red mat. \\
Move Stapler Pad & Pick up the stapler and place it on the red mat. \\
Stamp Seal & Grab the stamp and stamp onto the red mat. \\
Stack 3 Blocks & There are three blocks on the table, the color of the blocks is yellow, green and blue; move the blocks to the center of the table, and stack the blue block on the green block, and the green block on the yellow block. \\
Click 3 Bells & Click the right/left/middle bell. \\
\midrule
\multicolumn{2}{@{}l}{\textit{\textbf{Real World Tasks}}} \\
\midrule
Click 1 Bell & Click the bell on the table. \\
Click 2 Bells & Click the bell on the left/right. \\
Pick \& Place 1/3 Cup & Pick up the blue/green/yellow cup and place it on the coaster. \\
Pick \& Place 1/3 Block & Pick up the blue/green/red block and place it on the plate. \\
\bottomrule
\end{tabular}
\end{table*}

\clearpage

\end{document}